\documentclass[runningheads]{llncs}
\usepackage{graphicx}
\usepackage{comment}
\usepackage{amsmath,amssymb} 
\usepackage{color}

\usepackage{makecell}
\usepackage{cuted}
\usepackage{pifont}
\usepackage[dvipsnames]{xcolor}
\usepackage{balance}
\usepackage{epigraph}
\usepackage[misc]{ifsym}

\usepackage{tabularx}
\newcolumntype{Y}{>{\centering\arraybackslash}X}

\newcommand{\etc}{\textit{etc}}
\newcommand{\etal}{\textit{et al}}
\newcommand{\ie}{\textit{i.e.}}
\newcommand{\eg}{\textit{e.g.}}
\newcommand{\wrt}{\textit{w.r.t.} }

\usepackage[width=122mm,left=12mm,paperwidth=146mm,height=193mm,top=12mm,paperheight=217mm]{geometry}

\begin{document}
\pagestyle{headings}
\mainmatter
\def\ECCVSubNumber{22}  

\title{TSIT: A Simple and Versatile Framework for Image-to-Image Translation\\
\vspace{-0.2cm}
} 

\titlerunning{TSIT: A Simple and Versatile Framework for Image-to-Image Translation}
%
\author{Liming Jiang\inst{1} \and
Changxu Zhang\inst{2} \and
Mingyang Huang\inst{3} \and
Chunxiao Liu\inst{3} \and \\
Jianping Shi\inst{3} \and
Chen Change Loy\inst{1}$^{\textrm{\Letter}}$\\
\vspace{-0.15cm}
}
\authorrunning{L. Jiang et al.}
%
\institute{Nanyang Technological University \and
University of California, Berkeley \and
SenseTime Research \\
\email{liming002@ntu.edu.sg} \quad
\email{zhangcx@berkeley.edu} \\
\email{\{huangmingyang,liuchunxiao,shijianping\}@sensetime.com} \quad
\email{ccloy@ntu.edu.sg}
}
\maketitle



\begin{figure}
   \vspace{-0.55cm}
   \centering
       \includegraphics[width=\linewidth]{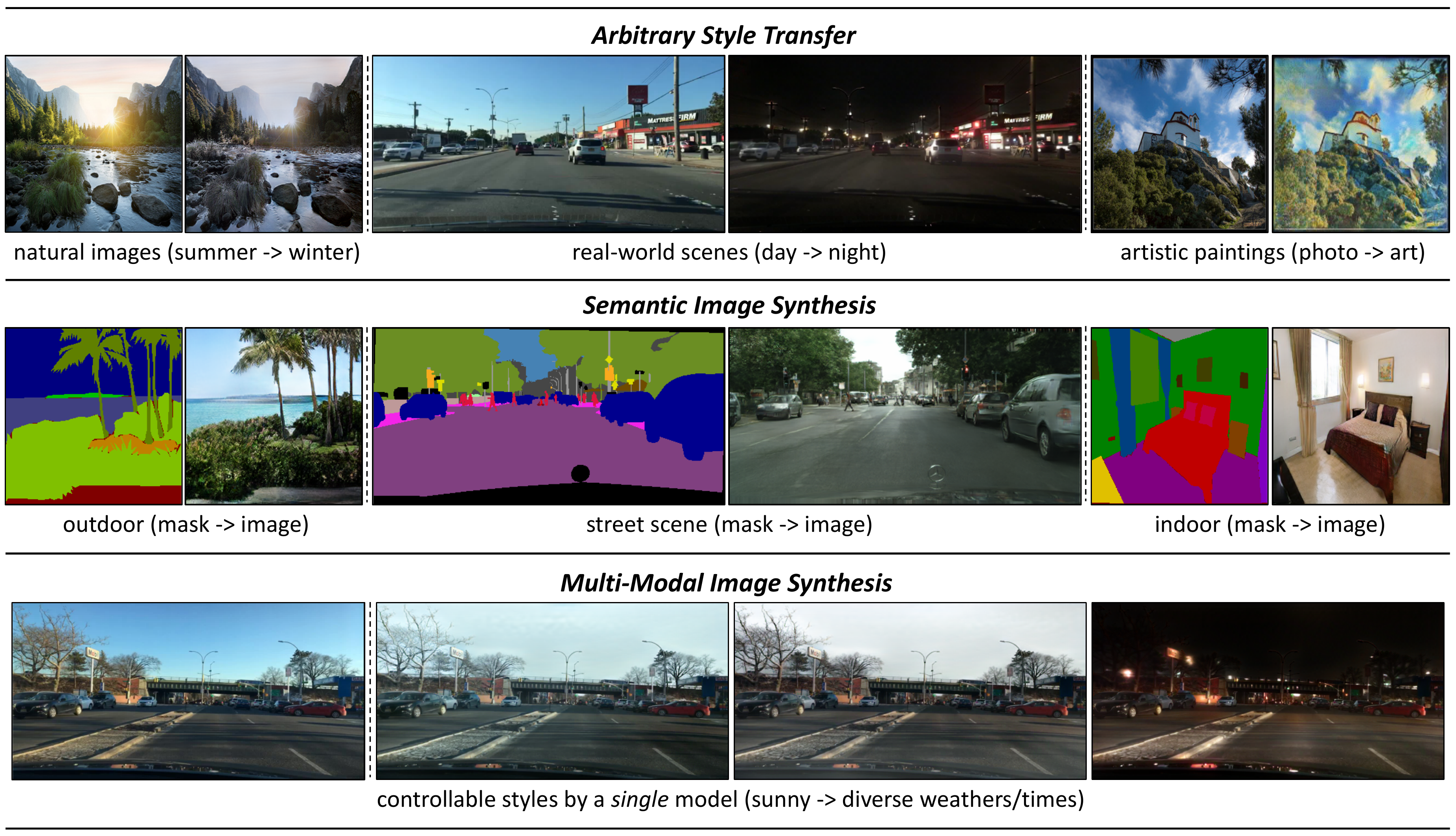}
   \vspace{-0.75cm}
      \caption{Our framework is simple and versatile for various image-to-image translation tasks. For unsupervised arbitrary style transfer,  diverse scenarios (\eg, natural images, real-world scenes, artistic paintings) can be handled. For supervised semantic image synthesis, our method is robust to different scenes (\eg, outdoor, street scene, indoor). Multi-modal image synthesis is feasible by a \textit{single} model with controllable styles.}
   \label{fig:teaser}
   \vspace{-0.85cm}
\end{figure}

\begin{abstract}
We introduce a simple and versatile framework for image-to-image translation. We unearth the importance of normalization layers, and provide a carefully designed two-stream generative model with newly proposed feature transformations in a coarse-to-fine fashion. This allows multi-scale semantic structure information and style representation to be effectively captured and fused by the network, permitting our method to scale to various tasks in both unsupervised and supervised settings. No additional constraints (\eg, cycle consistency) are needed, contributing to a very clean and simple method. Multi-modal image synthesis with arbitrary style control is made possible. A systematic study compares the proposed method with several state-of-the-art task-specific baselines, verifying its effectiveness in both perceptual quality and quantitative evaluations. GitHub: \textcolor{red}{\url{https://github.com/EndlessSora/TSIT}}.


\end{abstract}

\section{Introduction}
\label{introduction}


Image-to-image translation \cite{pix2pix} aims at translating one image representation to another. 
Recent advances \cite{GAN,congan,vae,convae,pixelcnn}, especially Generative Adversarial Networks (GANs) \cite{GAN}, have made remarkable success in various image-to-image translation tasks. 
Previous studies usually present specialized solutions for a specific form of application, ranging from arbitrary style transfer \cite{cyclegan,dualgan,adain,UNIT,MUNIT,DRIT,DMIT} in the unsupervised setting, to semantic image synthesis \cite{pix2pix,CRN,SIMS,pix2pixhd,SPADE,CC-FPSE} in the supervised setting.

In this study, we are interested in devising a general and unified framework that is applicable to different image-to-image translation tasks without degradation in synthesis quality.
This is \textit{non-trivial} given the different natures of different tasks. 
For instance, in certain conditional image synthesis tasks (\eg, arbitrary style transfer), paired data are usually not available. Under this unsupervised setting, translation task demands additional constraints on cycle consistency \cite{cyclegan,dualgan,DiscoGAN,UNIT}, semantic features \cite{DTN}, pixel gradients \cite{pixelgrad}, or pixel values \cite{pixelvalue}.
In semantic image synthesis (\ie, translation from segmentation labels to images), training pairs are available. This task is more data-dependent and typically needs losses to minimize per-pixel distance between the generated sample and ground truth. In addition, specialized structures~\cite{CRN,pix2pixhd,SPADE,CC-FPSE} are required to maintain spatial coherence and resolution.
Due to the different needs, existing methods exploit their own specially designed components. It is difficult to cross-use these components or integrate them into a unified framework.

%
%
%

To address the aforementioned challenges, we propose a Two-Stream Image-to-image Translation (TSIT) framework, which is \textit{versatile} for various image-to-image translation tasks (see Fig. \ref{fig:teaser}). 
The framework is simple as it is based purely on feature transformation.
Unlike previous approaches \cite{SPADE,adain} that only consider either semantic structure or style representation, we factorize \textit{both} the structure and style in multi-scale \textit{feature levels} via a symmetrical \textit{two-stream} network. The two streams jointly influence the new image generation in a coarse-to-fine manner via a consistent feature transformation scheme.
Specifically, the content spatial structure is preserved by an element-wise feature adaptive denormalization (FADE) from the content stream, while the style information is exerted by feature adaptive instance normalization (FAdaIN) from the style stream. 
Standard loss functions such as adversarial loss and perceptual loss are used, without additional constraints like cycle consistency. The pipeline is applicable to both unsupervised and supervised settings, easing the preparation of data.
 

%
%

The \textbf{contributions} of our work are summarized as follows. We propose TSIT, a simple and versatile framework, which is effective for various image-to-image translation tasks. Despite the succinct design, our network is readily adaptable to various tasks and achieves compelling results.
The good performance is achieved by 1) \textit{multi-scale} feature normalization (FADE and FAdaIN) scheme that captures \textit{coarse-to-fine} structure and style information, and 2) a \textit{two-stream} network design that integrates \textit{both} content and style effectively, reducing artifacts and making multi-modal image synthesis possible (see Fig. \ref{fig:teaser}). In comparison to several state-of-the-art task-specific baselines \cite{MUNIT,DMIT,CRN,SIMS,pix2pixhd,SPADE,CC-FPSE}, our method achieves comparable or even better results in both perceptual quality and quantitative evaluations.

\section{Related Work}
\label{relatedwork}


\noindent
\textbf{Image-to-image translation.}
%
Existing methods can be classified into two categories: unsupervised and supervised. 
%
With only unpaired data, unsupervised image-to-image translation problem is inherently ill-posed. Additional constraints are needed on \eg, cycle consistency \cite{cyclegan,dualgan,DiscoGAN,UNIT}, semantic features \cite{DTN}, pixel gradients \cite{pixelgrad}, or pixel values \cite{pixelvalue}.
In contrast, supervised methods, such as \texttt{pix2pix} \cite{pix2pix}, are more data-dependent, requiring well-annotated paired training samples. Subsequent approaches \cite{CRN,SIMS,pix2pixhd,SPADE,CC-FPSE} extend the supervised problem for generating high-resolution images or keeping effective semantic meaning.

Limited by learning only one-to-one mapping between two domains, some of the GAN-based methods \cite{cyclegan,dualgan,DiscoGAN,UNIT} suffer from generating images with low diversity. 
Recent studies explore more deeply into both multi-domain translation \cite{stargan,UFDN} and multi-modal translation \cite{MUNIT,DRIT,singlegan}, significantly increasing generation diversity. 
MUNIT \cite{MUNIT} is a representative method that disentangles the domain-invariant content and the domain-specific style representation, enriching the synthesized images. 
Multi-mapping translation is defined in a very recent work, DMIT \cite{DMIT}, which is designed to capture the multi-modal image nature in each domain.

Existing image-to-image translation methods lack the scalability to adapt to different tasks under diverse difficult settings. 
Different demands of unsupervised and supervised settings oblige previous methods to exploit customized modules. Cross-using these components will be suboptimal due to either degradation in quality or introduction of additional constraints. It is non-trivial to integrate them into a single framework and improve robustness.
In this study, we design a two-stream network with newly proposed feature transformations inspired by \cite{SPADE} and \cite{adain}. Our method is succinct yet able to link various tasks. 

\vspace{0.1cm}
\noindent
\textbf{Arbitrary style transfer.} Style transfer is closely relevant to image-to-image translation in the unsupervised setting. Style transfer aims at retaining the content structure of an image, while manipulating its style representation adopted from other images. 
Classical methods \cite{gatys,perceptualloss,stylebank,conIN} gradually improve this task from optimization-based to real-time, allowing multiple style transfer during inference.
%
Huang \etal. introduce AdaIN \cite{adain}, an effective normalization strategy for arbitrary style transfer. Several studies \cite{wavelettransformsst,graphcutst,textst,wctust,constydis,closedsolust,ETNet} improve stylization via wavelet transforms \cite{wavelettransformsst}, graph cuts \cite{graphcutst}, or iterative error-correction \cite{ETNet}. Besides, most collection-guided \cite{MUNIT} style transfer methods are GAN-based \cite{cyclegan,dualgan,UNIT,MUNIT,DRIT,DMIT}, showing impressive results.

Previous works usually consider either content or style information. In contrast, our framework succeeds in seeking a balance between content and style, and adaptively fuses them well. The proposed method achieves user-controllable multi-modal style manipulation by only a \textit{single} model. 
Compared to customized style transfer methods, our approach achieves better synthesis quality in many scenarios including natural images, real-world scenes, and artistic paintings.

\vspace{0.1cm}
\noindent
\textbf{Semantic image synthesis.} We define semantic image synthesis as in \cite{SPADE}, aiming at synthesizing a photorealistic image from a semantic segmentation mask. Semantic image synthesis is a special form of supervised image-to-image translation. The domain gap of this task is large. Therefore, keeping effective semantic information to enhance fidelity without losing diversity is challenging.

\texttt{Pix2pix} \cite{pix2pix} first adopts conditional GAN \cite{congan} in the semantic image synthesis task. 
Pix2pixHD \cite{pix2pixhd} contains a multi-scale generator and multi-scale discriminators to generate high-resolution images. 
SPADE \cite{SPADE} takes a noise map as input, and resizes the semantic label map for modulating the activations in normalization layers by a learned affine transformation. 
%
CC-FPSE \cite{CC-FPSE} employs a weight prediction network for generator. A semantics-embedding discriminator is used to enhance fine details and semantic alignments between the generated samples and the input semantic layouts.
In addition to these GAN-based methods, CRN \cite{CRN} applies a cascaded refinement network with regression loss as the supervision. 
SIMS \cite{SIMS} is a semi-parametric method, retrieving fragments from a memory bank and refining the canvas by a refinement network.

Different from prior works, we design a symmetrical two-stream framework. The network learns feature-level semantic structure information and style representation instead of directly resizing the input mask like SPADE \cite{SPADE}. Coarse-to-fine feature representations are learned by neural networks, adaptively keeping high fidelity without diminishing diversity. 
%


\vspace{0.25cm}
\section{Methodology}
\label{method}
We consider three key requirements in formulating a robust and scalable method to link various tasks:
1) \textit{Both} semantic structure information and style representation should be considered and fused adaptively. 
2) The content and style information should be learned by networks in \textit{feature level} instead of in image level to fit the nature of diverse semantic tasks. 
3) The network structure and loss functions should be \textit{simple} for easy training without additional constraints.

\begin{figure}[t]
	\centering
       \includegraphics[width=\linewidth]{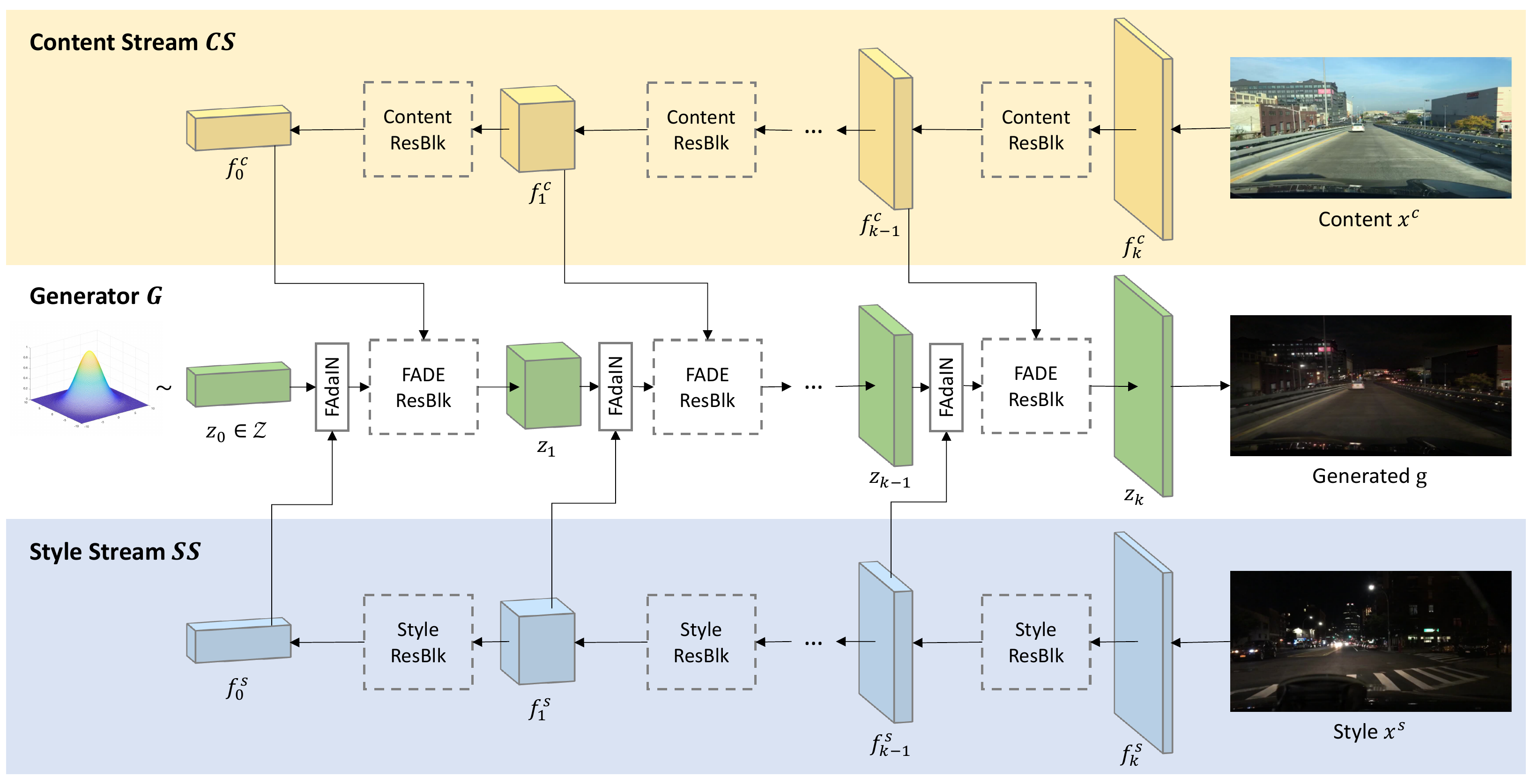}
      \caption{The proposed Two-Stream Image-to-image Translation (TSIT) framework. The multi-scale patch-based discriminators are omitted. A Gaussian noise map is taken as the latent input for the generator. The feature representations of the content and style images are extracted by the corresponding streams for multi-scale feature transformations. The symmetrical networks fuses semantic structure and style representation in an end-to-end training. Submodules of our network are shown in Fig. \ref{fig:sub_modules}.}
   \label{fig:main_framework}
\end{figure}

Based on the aforementioned considerations, we design a Two-Stream Image-to-image Translation (TSIT) framework (see Fig. \ref{fig:main_framework}).
We will detail our method in this section, including the network structure (Sec. \ref{networkstructure}), the feature transformation scheme (Sec. \ref{featuretransformation}), and the objective functions (Sec. \ref{objective}).

\begin{figure}[t]
	\centering
       \includegraphics[width=\linewidth]{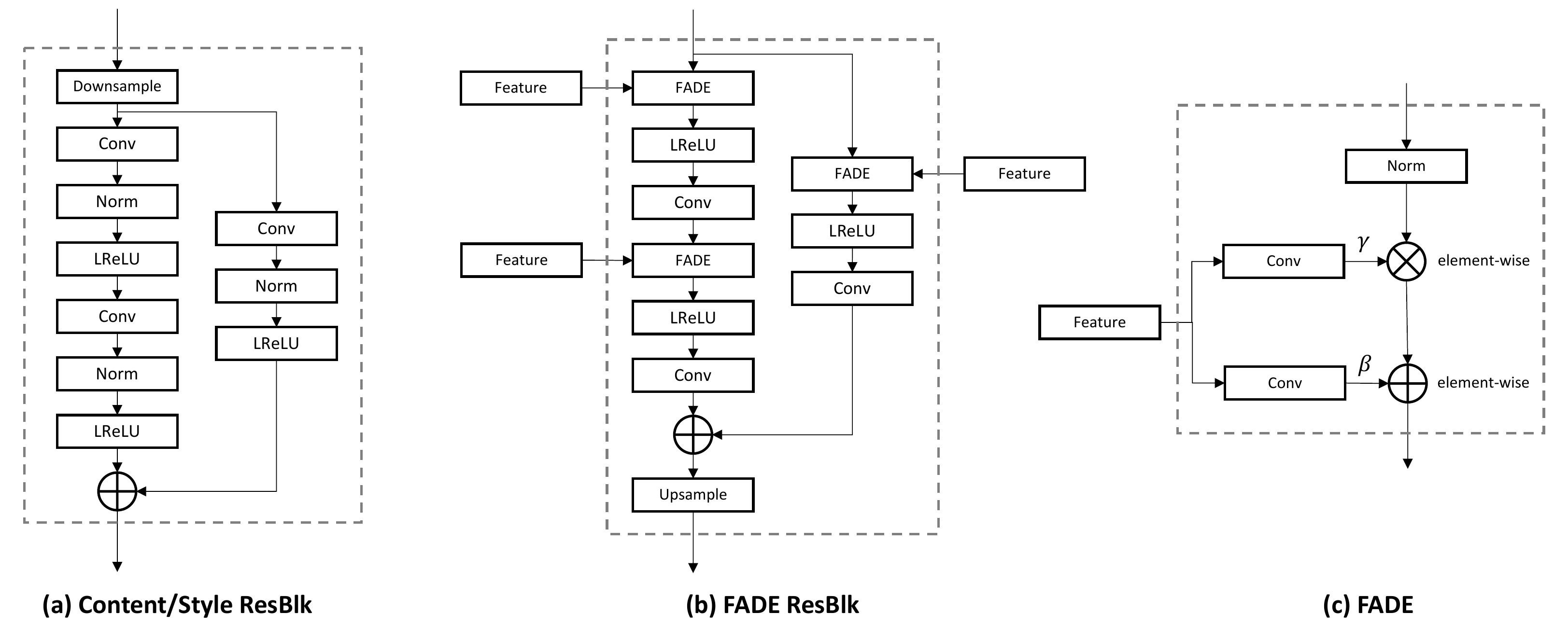}
      \caption{Submodules of our framework. (a) is a content/style residual block in the symmetrical content/style streams. (b) is a FADE residual block in the generator. (c) is a FADE module in the FADE residual block. It performs \textit{element-wise} denormalization by modulating the normalized activation using a learned affine transformation defined by the modulation parameters $\gamma$ and $\beta$.}
   \label{fig:sub_modules}
\end{figure}

\vspace{0.03cm}
\subsection{Network Structure}
\label{networkstructure}
As illustrated in Fig. \ref{fig:main_framework}, TSIT consists of four components: content stream, style stream, generator, and discriminators (omitted in Fig. \ref{fig:main_framework}). The first three main components are fully convolutional and symmetrically designed. The details of the submodules, including content/style residual block, FADE residual block, FADE module in the FADE residual block, are as shown in Fig. \ref{fig:sub_modules}. We will discuss them separately in this section.

\vspace{0.1cm}
\noindent
\textbf{Content/style stream.}
Unlike the traditional conditional GAN \cite{congan}, we place the two-stream networks, \ie, content stream and style stream, on each side of the generator (see Fig. \ref{fig:main_framework}). These two streams are symmetrical with the same network structure, aiming at extracting corresponding feature representations in different levels.
We construct content/style stream based on standard residual blocks \cite{resnet}. We call them content/style residual blocks. As shown in Fig \ref{fig:sub_modules} (a), each block has three convolutional layers, one of which is designed for the learned skip connection. The activation function is Leaky ReLU.
%
The function of content/style stream is to extract features and feed them to the corresponding feature transformation layers in the generator. Multi-scale content/style representation in \textit{feature levels} can be learned by content/style stream, adaptively fitting different feature transformations.

\vspace{0.1cm}
\noindent
\textbf{Generator.} The generator has a completely inverse structure \wrt the content/style stream. This is intentionally designed to consistently match the level of semantic abstraction at different feature scales.
A noise map is sampled from a Gaussian distribution as the latent input, and the feature maps from corresponding layers in content/style stream are taken as multi-scale feature inputs. The proposed feature transformations are implemented by a FADE residual block (Fig. \ref{fig:sub_modules} (b)) and a FAdaIN module. In the FADE residual block, we use an inverse architecture \wrt the content/style residual block and replace the batch normalization \cite{BN} layer with the FADE module (Fig. \ref{fig:sub_modules} (c)). The FADE module performs \textit{element-wise} denormalization by modulating the normalized activation using a learned affine transformation defined by the modulation parameters $\gamma$ and $\beta$. 
%
%
The FAdaIN module is used to exert style information through feature adaptive instance normalization. More discussions are given in Sec.~\ref{featuretransformation}.

The entire image generation process is performed in a coarse-to-fine manner. In particular, multi-scale content/style features are injected to refine the generated image constantly from high-level latent code to low-level image representation. Semantic structure and style information are learnable and effectively fused in an end-to-end training.

\vspace{0.1cm}
\noindent
\textbf{Discriminators.} We exploit the standard multi-scale patch-based discriminators (omitted in Fig. \ref{fig:main_framework}) in \cite{pix2pixhd,SPADE}. Three regular discriminators with an identical architecture are included to discriminate images at different scales. Despite the same structure, patch-based training allows the discriminator operating at the coarsest scale to have the largest receptive field, capturing global information of the image. Whereas the one operating at the finest scale has the smallest receptive field, making the generator produce better details. Multi-scale patch-based discriminators further improve the robustness of our method for image-to-image translation tasks in different resolutions.
Besides, the discriminators also serve as feature extractors for the generator to optimize the feature matching loss.

\subsection{Feature Transformation}
\label{featuretransformation}
We propose a new feature transformation scheme, considering \textit{both} semantic structure information and style representation, and fusing them adaptively. Let $x^c$ be the content image and $x^s$ be the style image. 
$CS$, $SS$, $G$, $D$ denote content stream, style stream, generator, and discriminators, respectively. 
Sampled from a Gaussian distribution, $z_0\in\mathbb{Z}$ is a noise map as the latent input for the generator (Fig. \ref{fig:main_framework}). Let $z_i\in\{z_0, z_1, z_2,...,z_k\}$ be the feature map after $i$-th residual block in the generator, with $k$ denoting the the total number of residual blocks (\ie, the upsampling times in the generator). Let $f_i^c\in\{f_0^c,f_1^c,f_2^c,...,f_k^c\}$ represent the corresponding feature representations extracted by the content stream (Fig. \ref{fig:main_framework}), $f_i^s\in\{f_0^s,f_1^s,f_2^s,...,f_k^s\}$ with the similar meaning in the style stream.

\vspace{0.1cm}
\noindent
\textbf{Feature adaptive denormalization (FADE).}
Our method is inspired by spatially adaptive denormalization (SPADE) \cite{SPADE}. Different from SPADE that resizes a semantic mask as its input, we generalize the input to multi-scale \textit{feature representation} $f_i^c$ of the content image $x^c$. 
In this way, we fully exploit semantic information captured by the content stream CS.

Formally, we define $N$ as the batch size, $L_i$ as the number of feature map channels in each layer. $H_i$ and $W_i$ are height and width, respectively. We first apply batch normalization \cite{BN} to normalize the generator feature map $z_i$ in a channel-wise manner. Then, we modulate the normalized feature by using the learned parameters scale $\gamma_i$ and bias $\beta_i$. The denormalized activation ($n\in{N}$, $l\in{L_i}$, $h\in{H_i}$, $w\in{W_i}$) is:
\begin{equation}
\label{eq:1}
\begin{split}
    \gamma_{i}^{l, h, w}\cdot \frac{z_i^{n, l, h, w}-\mu_{i}^{l}}{\sigma_{i}^{l}}+\beta_{i}^{l, h, w},
\end{split}
\end{equation}
where $\mu_{i}^{l}$ and $\sigma_{i}^{l}$ are the mean and standard deviation, respectively, of the generator feature map $z_i$ before the batch normalization \cite{BN} in channel $l$:
\begin{equation}
\label{eq:2}
\mu_{i}^{l}=\frac{1}{N H_{i} W_{i}}\sum_{n, h, w}z_i^{n, l, h, w},
\end{equation}
\begin{equation}
\label{eq:3}
\sigma_{i}^{l}=\sqrt{\frac{1}{N H_{i} W_{i}} \sum_{n, h, w}\left(z_{i}^{n, l, h, w}\right)^{2}-\left(\mu_{i}^{l}\right)^2}.
\end{equation}

The denormalization operation is \textit{element-wise}, and the parameters $\gamma^{l, h, w}_{i}$ and $\beta^{l, h, w}_{i}$ are learned by one-layer convolutions from $f_i^c$ in the FADE module (see Fig. \ref{fig:sub_modules} (c)). Compared to previous conditional normalization methods \cite{conIN,adain,SPADE}, FADE experiences more perceptible influence from coarse-to-fine feature representations, thus it can better preserve semantic structure information. 

\vspace{0.1cm}
\noindent
\textbf{Feature adaptive instance normalization (FAdaIN).} To better fuse style representation, we introduce another feature transformation, named feature adaptive instance normalization (FAdaIN). 
This method is inspired by adaptive instance normalization (AdaIN) \cite{adain}, with a generalization to enable the style stream $SS$ to learn multi-scale \textit{feature-level} style representation $f_i^s$ of the style image $x^s$ more effectively.

We use the same notation $z_i$ to represent the feature map after $i$-th FADE residual block in the generator. FAdaIN adaptively computes the affine parameters from the corresponding style feature $f_i^s$ with the same scale from $SS$:
\begin{equation}
\label{eq:4}
\begin{split}
    \mathrm{FAdaIN}\left(z_i, f_i^s\right)=\sigma\left(f_i^s\right)\left(\frac{z_i-\mu\left(z_i\right)}{\sigma\left(z_i\right)}\right)+\mu\left(f_i^s\right),
\end{split}
\end{equation}
where $\mu\left(z_i\right)$ and $\sigma\left(z_i\right)$ are the mean and standard deviation, respectively, of $z_i$.

Exploiting FAdaIN, coarse-to-fine style features at different layers can be fused adaptively with the corresponding semantic structure features learned by FADE, allowing our framework to be trained end-to-end and versatile to different tasks. 
Furthermore, owing to the effectiveness of FAdaIN in capturing multi-scale style feature representations, multi-modal image synthesis is made possible with arbitrary style control.

\subsection{Objective}
\label{objective}
We use standard losses in our objective function. Following \cite{SPADE,CC-FPSE}, we adopt a hinge loss term  \cite{geometricgan,spectralnorm,selfattentiongan} as our adversarial loss. For the generator, we apply hinge-based adversarial loss, perceptual loss \cite{perceptualloss}, and feature matching loss \cite{pix2pixhd}. For the multi-scale discriminators, only hinge-based adversarial loss is used to distinguish whether the image is real or fake. The generator and discriminator are trained alternately to play a min-max game. The generator loss $\mathcal{L}_{G}$ and the discriminator loss $\mathcal{L}_{D}$ can be written as:
\begin{equation}
\label{eq:5}
\begin{split}
    \mathcal{L}_{G}=
    -\mathbb{E}\left[D\left(g\right)\right]
    +\lambda_P\mathcal{L}_P\left(g,x^c\right)
    +\lambda_{FM}\mathcal{L}_{FM}\left(g,x^s\right),
\end{split}
\end{equation}
\begin{equation}
\label{eq:6}
\begin{split}
    \mathcal{L}_{D}=
    -\mathbb{E}\left[\min\left(-1+D\left(x^s\right),0\right)\right]
    -\mathbb{E}\left[\min\left(-1-D\left(g\right),0\right)\right],
\end{split}
\end{equation}
where $g=G\left(z_0,x^c,x^s\right)$ denotes the generated image, $z_0$, $x^c$, $x^s$ denote the input noise map in latent space, the content image, and the style image, respectively. $\mathcal{L}_{P}$ is the perceptual loss \cite{perceptualloss} that minimizes the difference between the feature representations extracted by VGG-19 \cite{perceptualloss} network. $\mathcal{L}_{FM}$ is the feature matching loss \cite{pix2pixhd} that matches the intermediate features at different layers of multi-scale discriminators. $\lambda_P$ and $\lambda_{FM}$ are the corresponding weights.

The simple objective functions make our framework stable and easy to train. Thanks to the two-stream network, the typical KL loss  \cite{vae,DMIT,SPADE,CC-FPSE} for multi-modal image synthesis becomes optional. Despite the simplicity, TSIT is a highly versatile tool, readily adaptable to various image-to-image translation tasks.




\vspace{0.2cm}
\section{Experiments}

\subsection{Settings}
\label{settings}

\vspace{0.1cm}
\noindent
\textbf{Implementation details.}
\label{implementationdetails}
We use Adam \cite{adam} optimizer and set $\beta_1=0$, $\beta_2=0.9$. Two time-scale update rule \cite{TTUR} is applied, where the learning rates for the generator (including two streams) and the discriminators are $0.0001$ and $0.0004$, respectively. We exploit Spectral Norm \cite{spectralnorm} for all layers in our network. We adopt SyncBN and IN \cite{IN} for the generator and the multi-scale discriminators, respectively. For the perceptual loss \cite{perceptualloss}, we use the feature maps of \texttt{relu1\_1}, \texttt{relu2\_1}, \texttt{relu3\_1}, \texttt{relu4\_1}, \texttt{relu5\_1} layers from a pretrained VGG-19 \cite{vgg} model, with the weights [1/32, 1/16, 1/8, 1/4, 1]. For the feature matching loss \cite{pix2pixhd}, we select features of three layers from the discriminator at each scale. All the experiments are conducted on NVIDIA Tesla V100 GPUs. Please refer to our \textit{Appendix} for additional implementation details.


\vspace{0.1cm}
\noindent
\textbf{Applications.}
\label{applicationexploration}
The proposed framework is versatile for various image-to-image translation tasks. We consider three representative applications of conditional image synthesis: arbitrary style transfer (unsupervised), semantic image synthesis (supervised), and multi-modal image synthesis (enriching generation diversity). Please refer to our \textit{Appendix} for details of our application exploration.

\vspace{0.1cm}
\noindent
\textbf{Datasets.}
\label{datasets}
For arbitrary style transfer, we consider diverse scenarios. We use Yosemite summer $\rightarrow$ winter dataset (natural images) provided by \cite{cyclegan}. We classify BDD100K \cite{bdd100k} (real-world scenes) into different times and perform day $\rightarrow$ night translation. Besides, we use Photo $\rightarrow$ art dataset (artistic paintings) in \cite{cyclegan}. For semantic image synthesis, we select several challenging datasets (\ie, Cityscapes \cite{cityscapes} and ADE20K \cite{ade20k}). For multi-modal image synthesis, we further classify BDD100K \cite{bdd100k} into different time and weather conditions, and perform controllable time and weather translation. The details of the datasets can be found in the \textit{Appendix}.

\vspace{0.1cm}
\noindent
\textbf{Evaluation metrics.}
\label{evaluationmetrics}
Besides comparing perceptual quality, we employ the standard evaluation protocol in prior works \cite{MUNIT,BigGAN,StyleGAN,SPADE,CC-FPSE} for quantitative evaluation. For arbitrary style transfer, we apply Fr$\mathrm{\acute{e}}$chet Inception Distance (FID, evaluating similarity of distribution between the generated images and the real images, lower is better) \cite{TTUR} and Inception Score (IS, considering clarity and diversity, higher is better) \cite{is}. For semantic image synthesis, we strictly follow \cite{SPADE,CC-FPSE}, adopting FID \cite{TTUR} and segmentation accuracy (mean Intersection-over-Union (mIoU) and pixel accuracy (accu)). The segmentation models are: DRN-D-105 \cite{drn} for Cityscapes \cite{cityscapes}, and UperNet101 \cite{upernet} for ADE20K \cite{ade20k}.

\vspace{0.1cm}
\noindent
\textbf{Baselines.}
\label{Baselines}
We compare our method with several state-of-the-art task-specific baselines. For a fair comparison, we mainly employ GAN-based methods. In the unsupervised setting, MUNIT \cite{MUNIT} and DMIT \cite{DMIT} are included, with the strong ability to capture the multi-modal nature of images while keeping quality. In the supervised setting, we compare against CRN \cite{CRN}, SIMS \cite{SIMS}, pix2pixHD \cite{pix2pixhd}, SPADE \cite{SPADE}, and CC-FPSE \cite{CC-FPSE}. 

\vspace{0.1cm}
\subsection{Results}
\label{results}

\begin{figure}[t]
	\centering
       \includegraphics[width=0.9\linewidth]{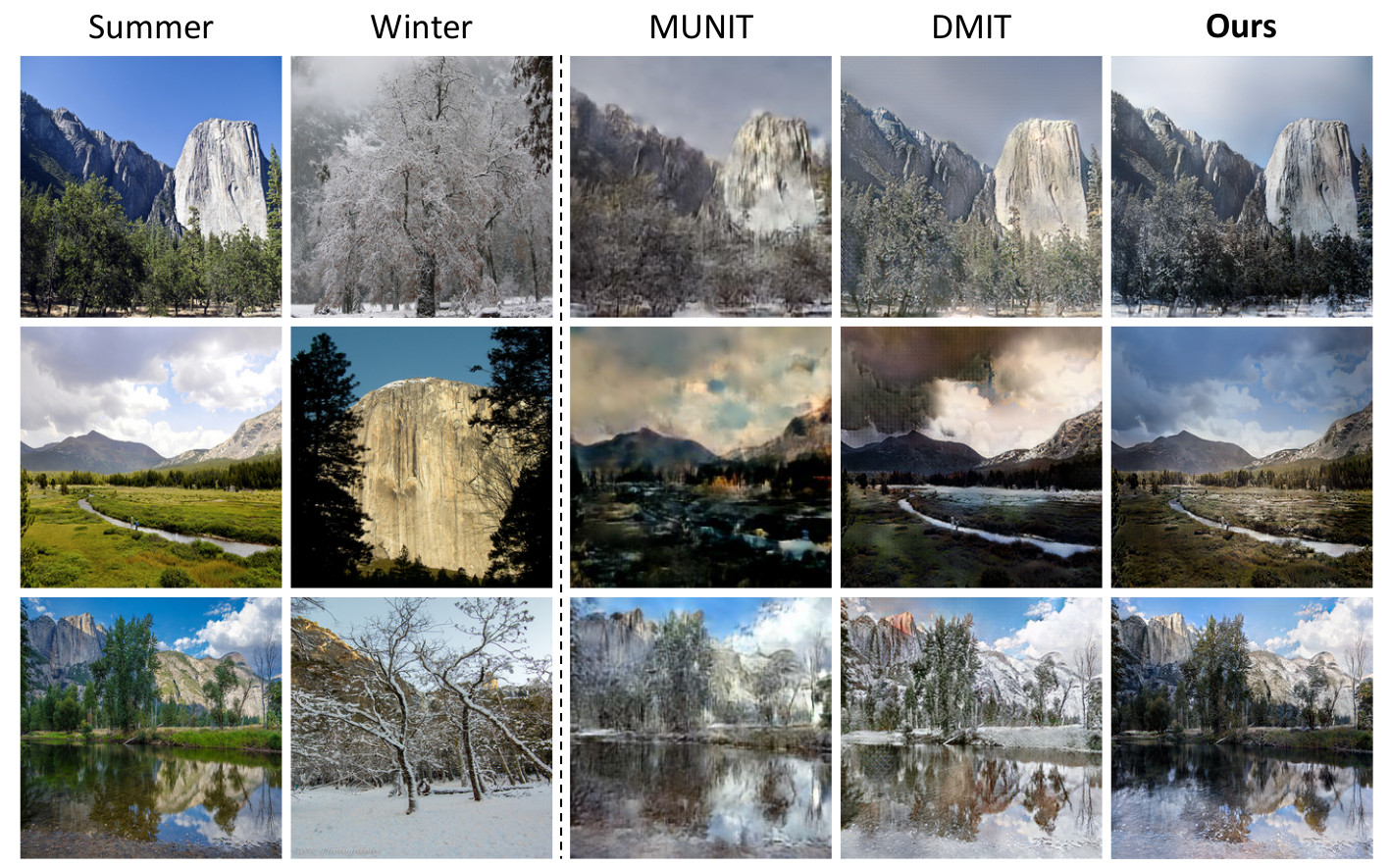}
   \vspace{-0.25cm}
      \caption{\textbf{Yosemite summer $\rightarrow$ winter} season transfer results compared to baselines.}
   \label{fig:season_compare}
   \vspace{-0.1cm}
\end{figure}

\begin{figure}[t]
	\centering
       \includegraphics[width=0.95\linewidth]{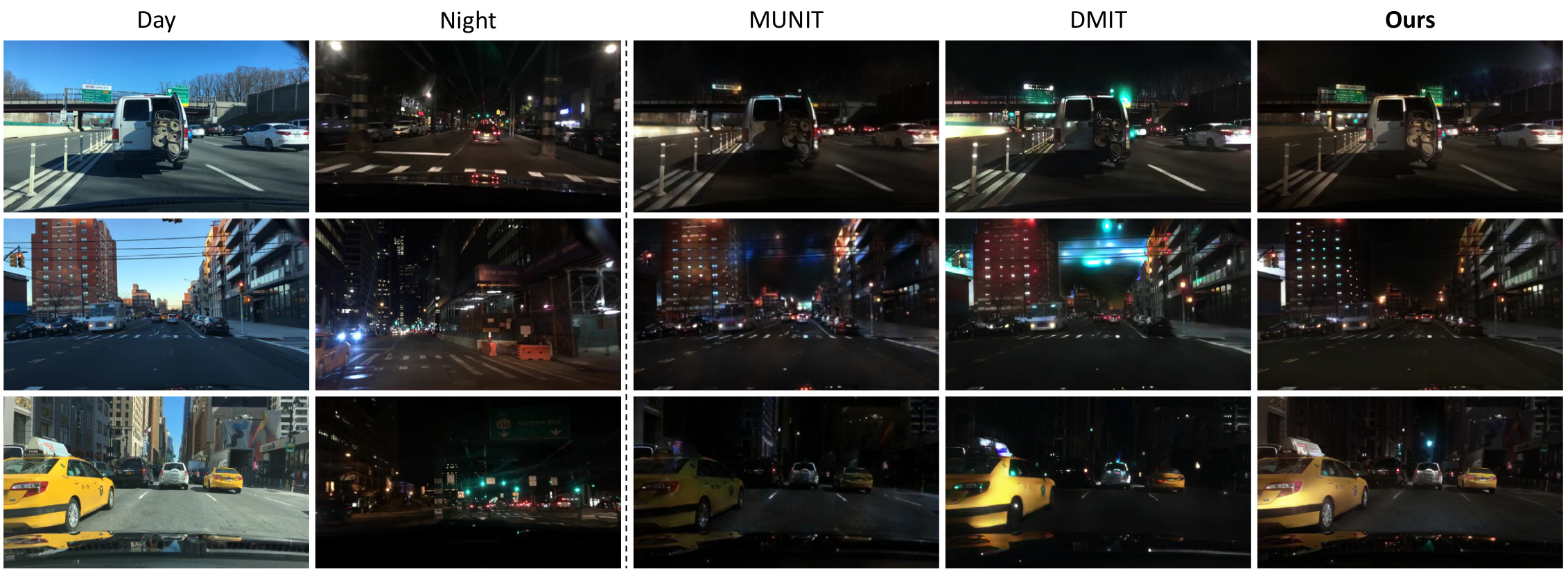}
   \vspace{-0.15cm}
      \caption{\textbf{BDD100K day $\rightarrow$ night} time translation results compared to baselines.}
   \label{fig:bdds2n_compare}
   \vspace{-0.1cm}
\end{figure}

\begin{figure}[t]
	\centering
       \includegraphics[width=0.9\linewidth]{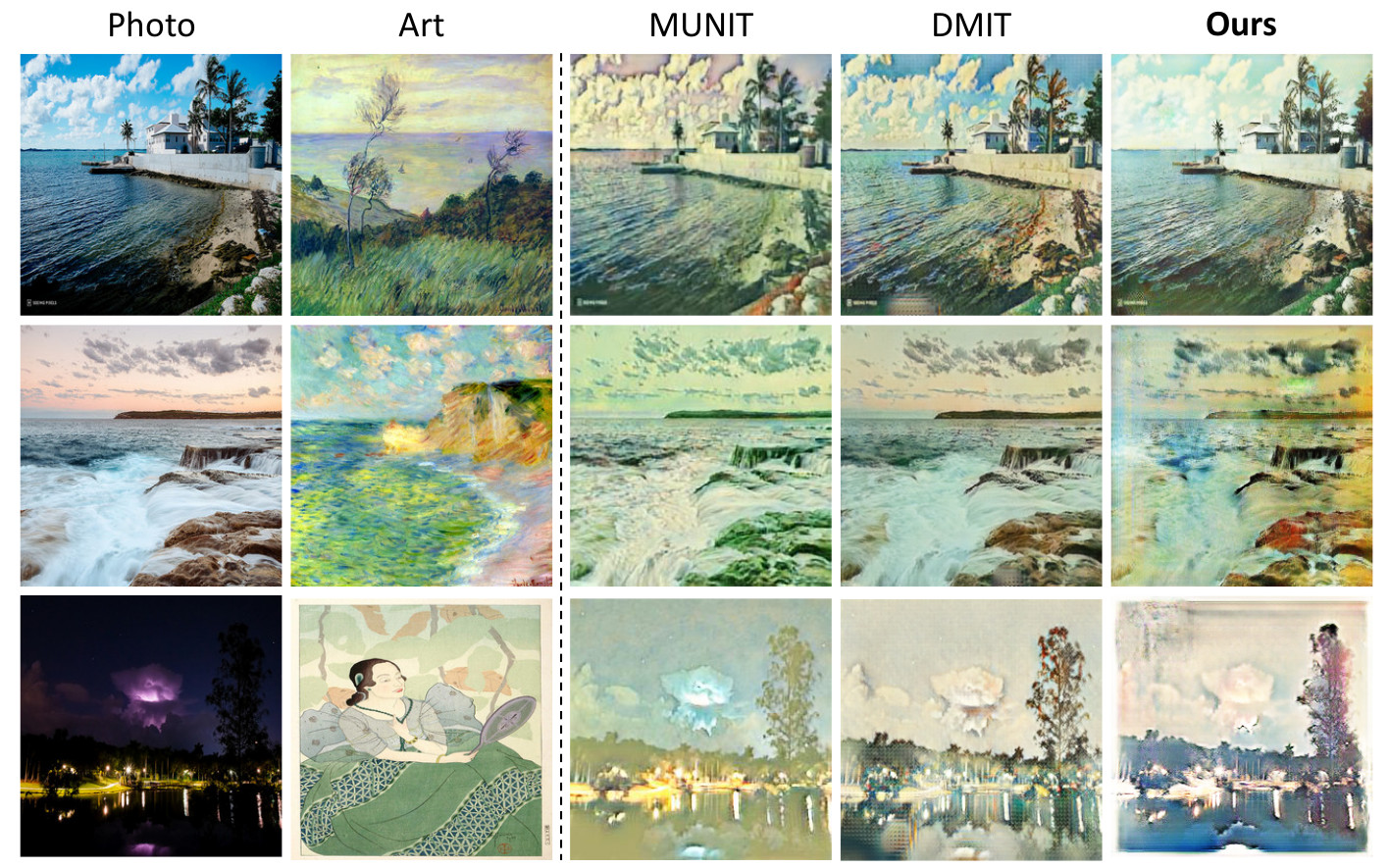}
   \vspace{-0.35cm}
      \caption{\textbf{Photo $\rightarrow$ art} style transfer results compared to baselines.}
   \label{fig:art_compare}
\end{figure}

\noindent
\textbf{Arbitrary style transfer.}
The results of \textit{Yosemite summer $\rightarrow$ winter season transfer} are shown in Fig. \ref{fig:season_compare}. Baselines \cite{MUNIT,DMIT} tend to impose the color of the style image (winter) to the whole content image (summer). Besides, MUNIT sometimes introduces unnecessary artistic effects, and DMIT generates some grid-like artifacts. In comparison, our generated results are clearer and more semantics-aware spatially. 
The results of \textit{BDD100K day $\rightarrow$ night time translation} are shown in Fig. \ref{fig:bdds2n_compare}. Some objects (\eg, road sign, car) generated by MUNIT are too dark, and the whole image tends to have some unnatural colors. DMIT introduces obvious artifacts to the car or sky. In contrast, our method produces more photorealistic samples in this task. 
In \textit{photo $\rightarrow$ art style transfer}, we choose some hard cases to make a clear comparison (see Fig. \ref{fig:art_compare}) due to the very strong ability of all the methods in this task. Our method can transfer the styles well while effectively keeping the content structure. MUNIT tends to impose a homogeneous color to the image. Although DMIT achieves slightly better stylization than our method in certain cases (in Row $3$ of Fig. \ref{fig:art_compare}), it also brings some grid-like distortions.

\begin{table}[tb!]
	\centering
	\caption{The FID and IS scores of our method compared to state-of-the-art methods in arbitrary style transfer tasks. A lower FID and a higher IS indicate better performance.}
	\vspace{-0.1cm}
    \begin{tabularx}{\textwidth}{l|*{6}{|Y}}
	\Xhline{1pt}
& \multicolumn{2}{c}{summer $\rightarrow$ winter} & \multicolumn{2}{|c}{day $\rightarrow$ night} & \multicolumn{2}{|c}{photo $\rightarrow$ art} \\
\cline{2-7}
Methods & FID $\downarrow$& IS $\uparrow$& FID $\downarrow$& IS $\uparrow$& FID $\downarrow$& IS $\uparrow$ \\
\Xhline{0.7pt}
MUNIT~\cite{MUNIT} & 118.225& 2.537& 110.011& 2.185& 167.314& 3.961 \\
DMIT~\cite{DMIT} & 87.969& 2.884& 83.898& 2.156& 166.933& 3.871 \\
Ours & {\bf80.138}& {\bf2.996}& {\bf79.697}& {\bf2.203}& {\bf165.561}& {\bf4.020} \\
	\Xhline{1pt}
	\end{tabularx}
	\label{tbl::ast_quantitative}		
 	\vspace{-0.45cm}
\end{table}

The quantitative evaluation results are shown in Table \ref{tbl::ast_quantitative}. Our approach achieves better performance than baselines \cite{MUNIT,DMIT} in all the tasks. 
We also note that the gap is relatively small in photo $\rightarrow$ art style transfer, in line with the close qualitative performance in this task (see Fig. \ref{fig:art_compare}).


\begin{figure}[t]
	\centering
       \includegraphics[width=0.92\linewidth]{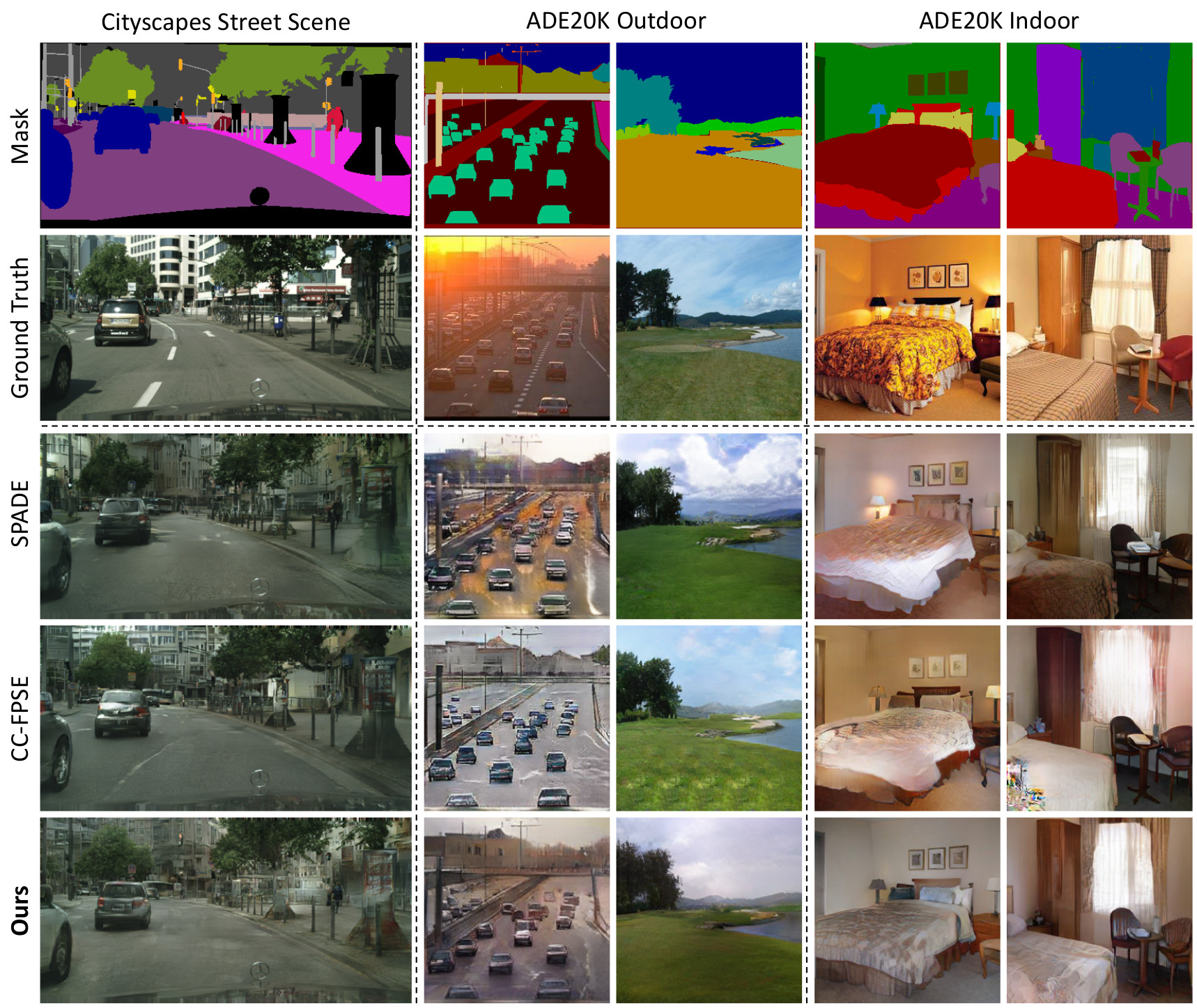}
   \vspace{-0.35cm}
      \caption{\textbf{Semantic image synthesis} results compared to baselines.}
   \label{fig:sis_compare}
\end{figure}

\vspace{0.1cm}
\noindent
\textbf{Semantic image synthesis.}
We choose two state-of-the-art baselines, SPADE \cite{SPADE} and CC-FPSE \cite{CC-FPSE}, to show some qualitative comparison results of semantic image synthesis (Fig. \ref{fig:sis_compare}). Our method demonstrates better perceptual quality than these task-specific baselines. 
In street scene (Column $1$), our method generates better details on key objects (car, pedestrian). 
In road scene (Column $2$), SPADE generates atypical colors on the roads, while CC-FPSE produces unnatural edges on the cars, hardly fitting the background (road). 
For outdoor natural images (Column $3$), all the methods share a similar generation quality. Our method is slightly better due to less distortions on the grass. 
In indoor scene (Column $4$ and $5$), SPADE and CC-FPSE produce obvious artifacts in some cases (Column $5$). In contrast, our method is more robust to diverse scenarios.

The quantitative evaluation results are shown in Table \ref{tbl::sis_quantitative} (the values used for comparison are taken from \cite{SPADE,CC-FPSE}). The proposed method achieves comparable performance with the very strong specialized methods \cite{CRN,SIMS,pix2pixhd,SPADE,CC-FPSE} for semantic image synthesis. Note that SIMS \cite{SIMS} yields the best FID score but poor segmentation performance on Cityscapes, because it stitches image patches from a memory bank of training set while not keeping the exactly consistent position in the synthesized image. Our approach achieves state-of-the-art segmentation performance on Cityscapes and the best FID score on ADE20K, suggesting its robustness to fit the nature of different image-to-image translation tasks.

\begin{table}[tb!]
	\vspace{-0.1cm}
	\centering
	\caption{The mIoU, pixel accuracy (accu) and FID scores of our method compared to state-of-the-art methods in semantic image synthesis tasks. A higher mIoU, a higher pixel accuracy (accu) and a lower FID indicate better performance.}
	\vspace{-0.15cm}
    \begin{tabularx}{\textwidth}{l|*{6}{|Y}}
	\Xhline{1pt}
& \multicolumn{3}{c}{Cityscapes} & \multicolumn{3}{|c}{ADE20K} \\
\cline{2-7}
Methods & mIoU $\uparrow$& accu $\uparrow$& FID $\downarrow$& mIoU $\uparrow$& accu $\uparrow$& FID $\downarrow$ \\
\Xhline{0.7pt}
CRN~\cite{CRN} & 52.4& 77.1& 104.7& 22.4& 68.8& 73.3 \\
SIMS~\cite{SIMS} & 47.2& 75.5& {\bf49.7} & N/A& N/A& N/A \\
pix2pixHD \cite{pix2pixhd} & 58.3& 81.4& 95.0& 20.3& 69.2& 81.8 \\
SPADE \cite{SPADE} &  62.3& 81.9& 71.8& 38.5& 79.9& 33.9 \\
CC-FPSE \cite{CC-FPSE} & 65.5& 82.3& 54.3& {\bf43.7}& {\bf82.9}& 31.7 \\
Ours &  {\bf65.9}& {\bf82.7}$^*$& 59.2& 38.6& 80.8& {\bf31.6} \\
	\Xhline{1pt}
	\end{tabularx}
	\label{tbl::sis_quantitative}		
 	\vspace{-0.45cm}
\end{table}
{\let\thefootnote\relax\footnotetext{$^*$ The value differs from the earlier version of this paper~\cite{tsitarxivv1}. The official code of DRN~\cite{drn} does not provide the implementation of the ``accu'' metric. The new accu value $82.7\%$ (still the best among the compared methods) is obtained by including $255$-labeled pixels, consistent with~\cite{SPADE,CC-FPSE}. The previously reported accu $94.4\%$ omits $255$-labeled pixels, which may be more reasonable due to its consistency with the training of the segmentation model and the calculation of mIoU.}}


\begin{figure}[t]
	\centering
       \includegraphics[width=\linewidth]{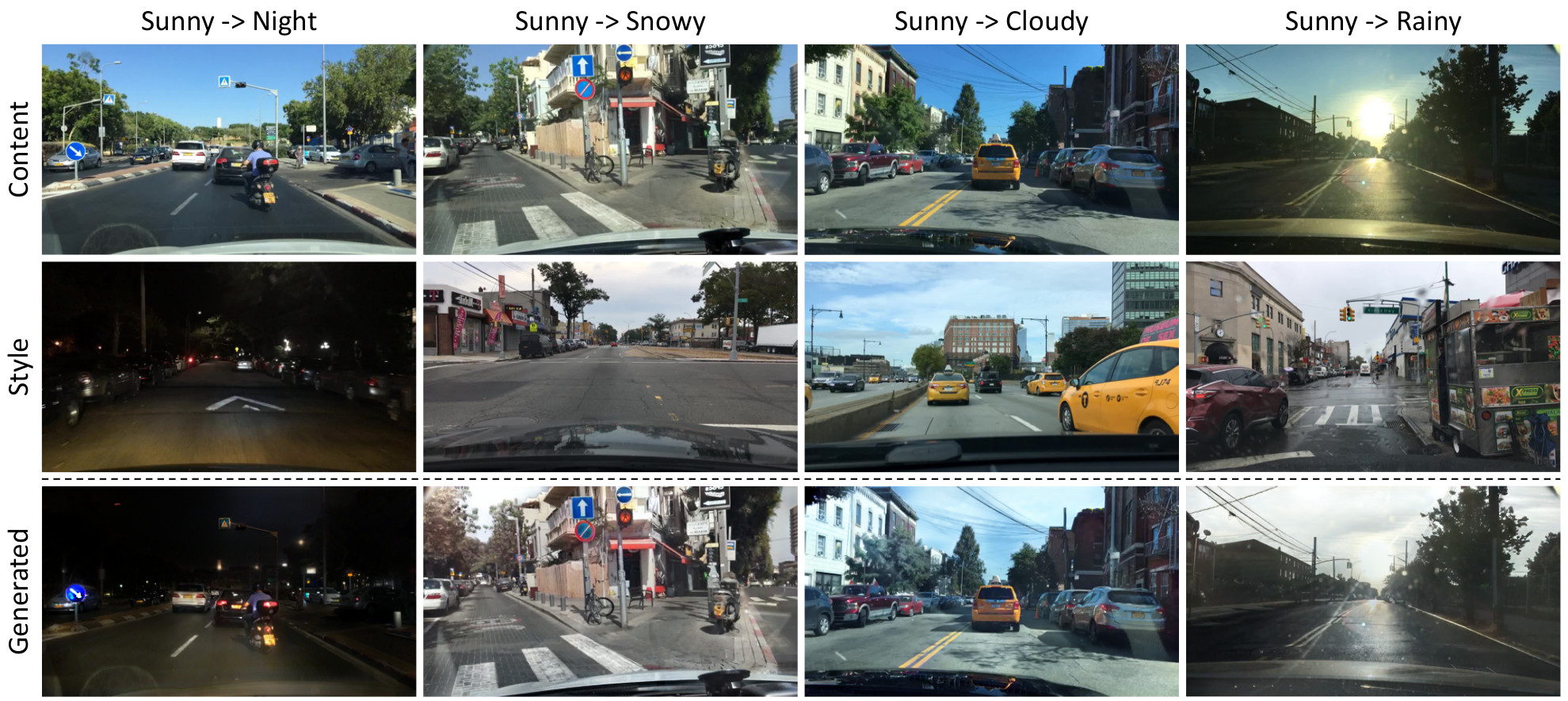}
      \caption{\textbf{BDD100K multi-modal image synthesis} results for different time and weather translation by a \textit{single} model.}
   \label{fig:multibdd}
\end{figure}

\begin{figure}[t]
	\centering
       \includegraphics[width=\linewidth]{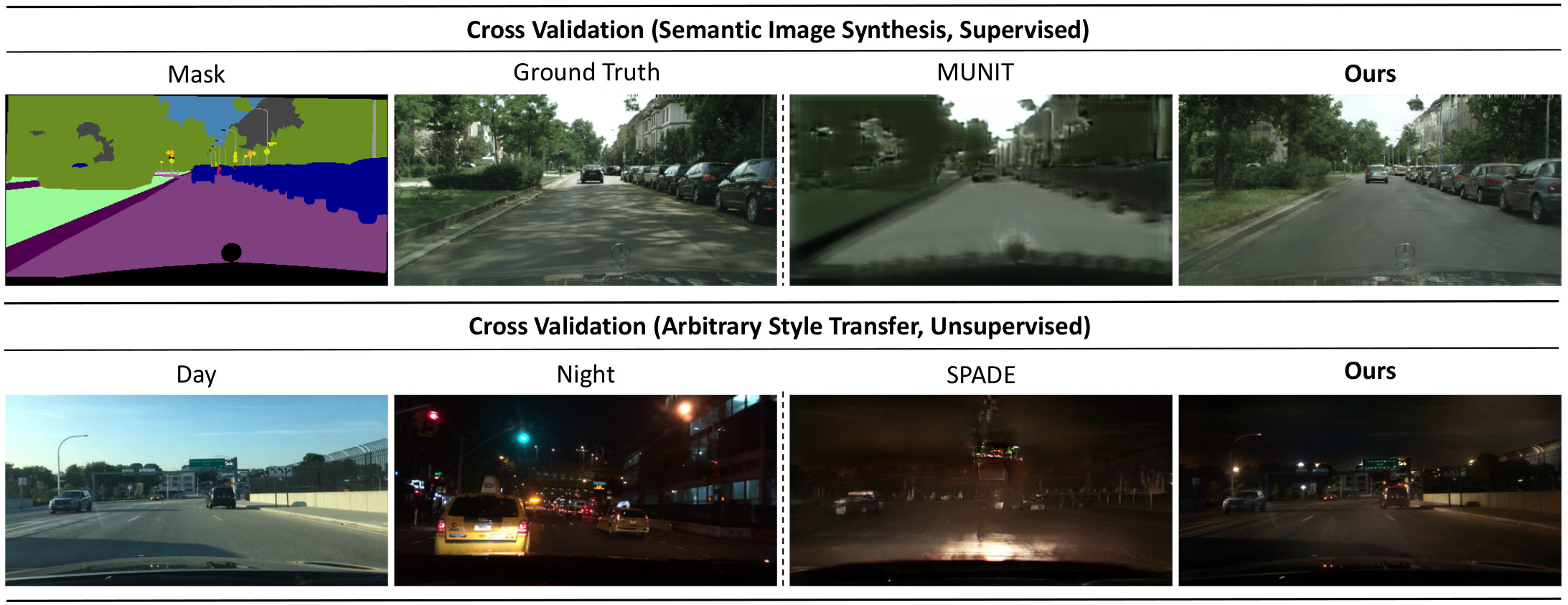}
      \caption{\textbf{Cross validation} of ineffectiveness of task-specific methods in inverse settings.}
   \label{fig:cross_val}
\end{figure}

\vspace{0.1cm}
\noindent
\textbf{Multi-modal image synthesis.}
We perform multi-modal image synthesis for time and weather image-to-image translation (see Fig. \ref{fig:multibdd}) on BDD100K \cite{bdd100k}. Training only a \textit{single} model, we translate the images of weather \textit{sunny} to different times and weathers (\ie, \textit{night}, \textit{snowy}, \textit{cloudy}, \textit{rainy}). Our method effectively adapts to different style control and keeps photorealistic generation quality. Although the weather \textit{snowy} is not very obvious in BDD100K \cite{bdd100k}, our approach successfully introduces some snow-like effects on trees and grounds (Column $2$).

\begin{figure}[t]
	\centering
       \includegraphics[width=0.85\linewidth]{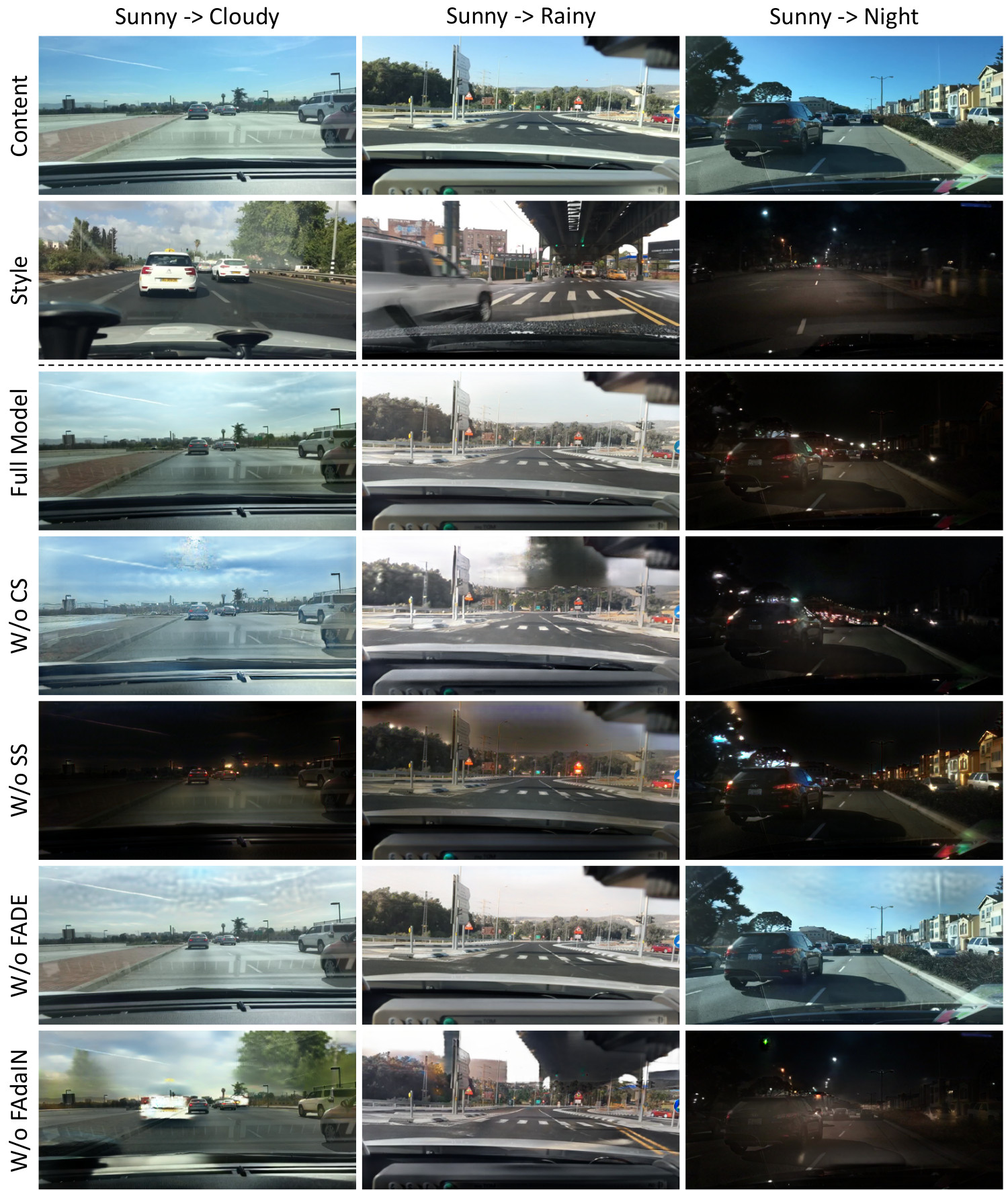}
      \caption{\textbf{Ablation studies} of key modules (\ie, content stream (CS), style stream(SS)) and feature transformations in multi-modal image synthesis task.}
   \label{fig:ablation}
\end{figure}

\vspace{0.1cm}
\noindent
\textbf{Cross validation.}
We also conduct experiments to evaluate the performance of existing specialized methods in inverse settings (\ie, using unsupervised methods to do semantic image synthesis / using supervised methods to perform arbitrary style transfer). We selected two representative methods, MUNIT \cite{MUNIT} and SPADE \cite{SPADE}. Without modifying the architecture, we tuned the loss weights and tried to get the best generation results. To ensure a fair comparison, we also tried to compute perceptual loss with the content (day) image for SPADE to match the setting of TSIT. Representative results of cross validation are shown in Fig. \ref{fig:cross_val}. The proposed method shows much better results than baseline methods. MUNIT fails to adapt to semantic image synthesis. SPADE loses details of key objects and introduces very strong artifacts despite translating the color correctly.

\vspace{0.1cm}
\noindent
\textbf{Ablation studies.}
We present ablation studies of key modules (\ie, content stream (CS), style stream(SS)) and the proposed feature transformations (see Fig. \ref{fig:ablation}. More ablation study results can be found in the \textit{Appendix}). We perform multi-modal image synthesis to show the effectiveness of different components.
Our full model generates high-quality results (Row $3$). 
When we directly inject the resized content image without CS, the semantic structure information becomes weak, leading to several artifacts in the sky (Row $4$). 
Without SS, the model cannot perform multi-modal image synthesis at all (Row $5$). The style representation is dominated by the night style. 
When we concatenate the feature maps of CS with the ones of the generator instead of using FADE, the concatenation introduces too much content information, leading to several failure cases (\eg, \textit{sunny $\rightarrow$ night} in Row $6$). 
If we discard FAdaIN by concatenating the feature maps of SS with the ones of the generator, the style becomes too strong, causing serious style regionalization problem (Row $7$).



\section{Conclusion}
We propose TSIT, a simple and versatile framework for image-to-image translation. The proposed symmetrical two-stream network allows the image generation to be effectively conditioned on the multi-scale feature-level semantic structure information and style representation via feature transformations. 
A systematic study verifies the effectiveness of our method in diverse tasks 
 compared to state-of-the-art task-specific baselines. 
We believe that designing a unified and versatile framework for more tasks is an important direction in the image generation area.
%
Incorporating unconditional image synthesis tasks and introducing more variability into the two streams/latent space can be interesting future works.

\vspace{0.1cm}
\noindent
\textbf{Acknowledgements.} This work is supported by the SenseTime-NTU Collaboration Project, Singapore MOE AcRF Tier 1 (2018-T1-002-056), and NTU NAP.


\bibliographystyle{splncs04}
\bibliography{sections_arxiv/egbib}


\clearpage
\section*{Appendix}
\label{appendix}
This appendix provides the supplementary information that is not elaborated in the main paper: Sec.~\ref{sec:appexpl} details the different applications we have explored. Sec.~\ref{sec:datasetdetails} describes details of the datasets used in our experiments. Sec.~\ref{sec:addiimpdetails} provides additional implementation details. Sec.~\ref{sec:suppablation} presents several supplementary ablation studies. Sec.~\ref{sec:suppexamples} shows more examples of the images generated by our method.

\appendix
\section{Application Exploration}
\label{sec:appexpl}
We have introduced a Two-Stream Image-to-image Translation (TSIT) framework in the main paper. The proposed framework is simple and versatile for various image-to-image translation tasks under both unsupervised and supervised settings. We have considered three important and representative applications of conditional image synthesis: arbitrary style transfer (unsupervised), semantic image synthesis (supervised), and multi-modal image synthesis (enriching generation diversity). We employ a two-stream network, namely ``content'' stream and ``style'' stream, on these applications.

For the unsupervised arbitrary style transfer application, we feed the content image to the content stream and the style image to the style stream, and let the networks learn different levels of \textit{feature representations} of the content and style. The proposed feature transformations, FADE and FAdaIN, adaptively fuse content and style feature maps, respectively, at different scales in the generator. In contrast to prior works, our method is more adaptable to style transfer tasks in diverse scenarios (\eg, natural images, real-world scenes, and artistic paintings).

We further expand the application of our method to cater to semantic image synthesis under the supervised setting. The definition of ``content'' and ``style'' can be more general: all the images that provide semantic structure information can be content images, and all the images representing the global style distribution can be considered as style images. Therefore, when we inject semantic segmentation masks to the content stream and the corresponding real images to the style stream, semantic image synthesis task in the supervised setting can be handled. Despite a rather large domain gap in this task, our framework yields comparable or even better results over the state-of-the-art task-specific methods, suggesting the high adaptability of our approach.

It is noteworthy to highlight that the newly proposed feature transformations and the symmetrical two-stream network can effectively disentangle the semantic structure and style information. Thanks to the clean disentanglement, the high-level multi-modal nature of the images can be captured by our framework, contributing to high-fidelity multi-modal image synthesis.

\section{Dataset Details}
\label{sec:datasetdetails}
In this section, we discuss the detailed information of all the datasets we explored, including the source, preprocessing, number of images, resolution, \etc.

For arbitrary style transfer under the unsupervised setting, paired data are not needed. We perform style transfer tasks in diverse scenarios (\eg, natural images, real-world scenes, and artistic paintings).
\begin{itemize}
	\item \textbf{Yosemite summer $\rightarrow$ winter.} We use this unpaired dataset provided by \cite{cyclegan}, containing rich natural images collected via Flickr API. We perform season transfer using this dataset, with $1,231$ summer images and $962$ winter images for training. The resolution is $256 \times 256$.
	\item \textbf{BDD100K day $\rightarrow$ night.} We conduct time translation on BDD100K \cite{bdd100k} dataset, which is captured at diverse locations in the United States. All the images are in real-world scenes, mostly street/road scenes. We classify the dataset into different times. The training set contains $12,454$ daytime images and $22,884$ nighttime images. The original images are scaled to $512 \times 256$.
	\item \textbf{Photo $\rightarrow$ art.} We utilize the art dataset collected in \cite{cyclegan}. The art images of this dataset were downloaded from Wikiart.org. The dataset consists of photographs and diverse artistic paintings (Monet: $1,074$; C$\mathrm{\acute{e}}$zanne: $584$; van Gogh: $401$; Ukiyo-e: $1,433$). To test the robustness of the models for arbitrary style transfer, we combine all the artistic styles, yielding $6,853$ photos and $3,492$ paintings for training. All the images are uniformly resized to $256 \times 256$.
\end{itemize}

For semantic image synthesis under the supervised setting, we follow \cite{CC-FPSE,SPADE} and select several challenging datasets.
\begin{itemize}
	\item \textbf{Cityscapes.} Cityscapes \cite{cityscapes} dataset contains street scene images mostly collected in Germany, with $2,975$ images for training and $500$ images for evaluation. The dataset provides instance-wise, dense pixel annotations of $30$ classes. All the image sizes are adjusted to $512 \times 256$.
	\item \textbf{ADE20K.} We use ADE20K \cite{ade20k} dataset consisting of challenging in-the-wild images with fine annotations of $150$ semantic classes. The sizes of training and validation sets are $20,210$ and $2,000$, respectively. All the images are scaled to $256 \times 256$.
\end{itemize}

For multi-modal image synthesis, we use BDD100K \cite{bdd100k} dataset, details of which have been described earlier.
\begin{itemize}
	\item \textbf{BDD100K sunny $\rightarrow$ different time/weather conditions}. We further classify the images in BDD100K \cite{bdd100k} dataset into different time and weather conditions, constituting a training set of $10,000$ sunny images and $10,000$ images of other time and weather conditions (night: $2,500$; cloudy: $2,500$; rainy: $2,500$; snowy: $2,500$). The resolution is $512 \times 256$.
\end{itemize}


\section{Additional Implementation Details}
\label{sec:addiimpdetails}
We provide more implementation details in this section, including the network architecture specifics, detailed feature shapes, hyperparameters, \etc.

\vspace{0.1cm}
\noindent
\textbf{Network architecture specifics.}
Our framework consists of four components: content stream, style stream, generator, and discriminators. The first three components maintain a symmetrical structure, using fully convolutional networks. The number of residual blocks $k$ (\ie, downsampling/upsampling times) in the content/style stream and the generator equals to $7$. Let $inc$, $outc$, $kn$, $s$, $p$ denote the input channel, the output channel, the kernel size, the stride, and the zero-padding amount, respectively.

In the content/style stream, we use a series of content/style residual blocks with the nearest neighbor downsampling. The scale factor of downsampling is $2$. By default, we use instance normalization \cite{IN} for the content/style residual blocks, and the negative slope of Leaky ReLU is $0.2$. Thus, the structure of Content/Style ResBlk$(inc,outc)$ is: $\mathrm{Downsample}(2)\mathrm{-Conv}(inc,inc,kn3\times3,s1,p1)\mathrm{-IN-LReLU}(0.2)\mathrm{-Conv}(inc,outc,kn3\times3,s1,p1)\mathrm{-IN-LReLU}(0.2)$ with the learned skip connection $\mathrm{Conv}(inc,outc,kn1\times1,s1,p0)\mathrm{-IN-LReLU}(0.2)$.

In the generator, we construct several FADE residual blocks with the nearest neighbor upsampling. The scale factor of upsampling is $2$. FAdaIN layers are applied before each FADE residual block. The FADE residual block contains a FADE submodule, which performs \textit{element-wise} denormalization using a learned affine transformation defined by the modulation parameters $\gamma$ and $\beta$. Let $normc$, $featc$ indicate the normalized channel and the injected feature channel, respectively. Then, the convolutional layers in FADE$(normc,featc)$ can be represented as: $\mathrm{Conv}(featc,normc,kn3\times3,s1,p1)$. By default, we adopt SyncBN for the generator, and the negative slope of Leaky ReLU is $0.2$. The structure of FADE ResBlk$(inc,outc)$ is: $\mathrm{FADE}(inc,inc)\mathrm{-LReLU}(0.2)\mathrm{-Conv}(inc,inc,kn3\times3,s1,p1)\mathrm{-FADE}(inc,inc)\mathrm{-LReLU}(0.2)\mathrm{-Conv}(inc,outc,kn3\times3,s1,p1)\mathrm{-Upsa}\text{-}\\\mathrm{mple}(2)$ with the learned skip connection $\mathrm{FADE}(inc,inc)\mathrm{-LReLU}(0.2)\mathrm{-Conv}(inc,\\outc,kn1\times1,s1,p0)$.

As mentioned in the main paper, we exploit the same multi-scale patch-based discriminators as \cite{pix2pixhd,SPADE}. The detailed network architectures and the layers used for feature matching loss \cite{pix2pixhd} are also identical.

\vspace{0.1cm}
\noindent
\textbf{Feature shapes.}
In the content/style stream, we put an input layer at the entrance. The feature channel is adjusted to $64$ after the input layer, while the resolution remains unchanged. Then, the feature channels after each of the $k (7)$ residual blocks are: $128$, $256$, $512$, $1024$, $1024$, $1024$, $1024$. Since the scale factor of downsampling is $2$ (as described in the network architecture specifics above), the resolution of the features is halved after each residual block.
The generator feature shapes are strictly corresponding and opposite to that of content/style stream.
The discriminator feature shapes are identical to that in \cite{pix2pixhd,SPADE}, where the resolution is halved on every step of the pyramid.

\vspace{0.1cm}
\noindent
\textbf{Additional training details.}
For perceptual loss, we use the feature reconstruction loss that requires a content target \cite{perceptualloss}. 

In the arbitrary style transfer and multi-modal image synthesis tasks, the content target is the content image. The loss weights are $\lambda_P=1,\lambda_{FM}=1$, and the batch size is $1$. We train our models for $200$ epochs on Yosemite summer $\rightarrow$ winter, $10$ epochs on BDD100K day $\rightarrow$ night, $40$ epochs on Photo $\rightarrow$ art, and $20$ epochs on BDD100K sunny $\rightarrow$ different time/weather conditions. The models are trained on $1$ NVIDIA Tesla V100 GPU, with around $10$ GB memory consumption. For multi-modal image synthesis, similar to \cite{pix2pix}, at the inference phase we run the generator network in exactly the same manner as during the training phase. For the cross validation of SPADE \cite{SPADE}, the hyperparameters obtaining the best generation results are $\lambda_P=10,\lambda_{FM}=10$.

In the semantic image synthesis task, the content target is the ground truth real image. The corresponding loss weights are $\lambda_P=20,\lambda_{FM}=10$, and the batch size is $16$. We perform $200$ epochs of training on Cityscapes and ADE20K. The models are trained on $2$ NVIDIA Tesla V100 GPUs, each with about $32$ GB memory consumption. We also find that in semantic image synthesis, weakening/removing the style stream can sometimes contribute to a performance boost. Besides, exploiting variational auto-encoders \cite{vae} can help in certain cases. For the cross validation of MUNIT \cite{MUNIT}, since the loss functions are very different from ours, we use its default hyperparameters in unsupervised image-to-image translation.


\section{Supplementary Ablation Studies}
\label{sec:suppablation}
We ablate the key modules (\ie, content stream (CS), style stream(SS)) and the proposed feature transformations in the main paper. We perform multi-modal image synthesis to clearly show the effectiveness of different components.
Due to the space constraints, we only provide qualitative evaluation results. In this section, we will first show the quantitative evaluation results of key component ablation studies in the main paper. Then, we will dig deeper and present more supplementary ablation study results.

\vspace{0.1cm}
\noindent
\textbf{Quantitative evaluation of key component ablation studies.}
We conduct quantitative evaluation on ablation studies of the key components in multi-modal image synthesis task.
As shown in Table~\ref{tbl:ablation_quantitative}, using the full model we introduced, the lowest FID score and highest IS score have been achieved. This means the generated images by our full model are the most photorealistic, clearest, and of the highest diversity. 
Without any key module of TSIT, the quantitative performance will drop. This verifies the necessity of these components for our method.

\begin{table}[tb!]
\centering
\caption{The quantitative evaluation on ablation studies of the key modules (\ie, content stream (CS), style stream (SS)) and the feature transformations in multi-modal image synthesis task. A lower FID and a higher IS indicate better performance.}
\begin{tabularx}{\textwidth}{l|*{5}{|Y}}
\Xhline{1pt}
& \multicolumn{5}{c}{multi-modal image synthesis} \\
\cline{2-6}
Metrics & full model& w/o CS& w/o SS& w/o FADE& w/o FAdaIN \\
\Xhline{0.7pt}
FID $\downarrow$& {\bf85.876}& 89.429& 86.263& 86.463& 89.795 \\
IS $\uparrow$& {\bf2.934}& 2.851& 2.734& 2.881& 2.890 \\
\Xhline{1pt}
\end{tabularx}
\label{tbl:ablation_quantitative}		
\end{table}

\vspace{0.1cm}
\noindent
\textbf{Feature channel ablation studies.}
We also study how the number of feature channels in the two streams (\ie, content stream (CS) and style stream (SS)) affects the image synthesis results. We conduct quantitative evaluation of feature channel ablation studies, covering all of the discussed tasks. Note that we should change the channels in CS/SS at the same time to maintain a symmetrical structure.
As presented in Table~\ref{tbl:ablation_channel_ast}, Table~\ref{tbl:ablation_channel_sis} and Table~\ref{tbl:ablation_channel_mmis}, in different tasks under either unsupervised or supervised setting, the best performance is achieved by the full model of TSIT. As we reduce the channel numbers in the two-stream network, the image synthesis quality gradually degrade.
For more channels, memory consumption will increase exponentially.

\begin{table}[tb!]
\centering
\caption{The quantitative evaluation on ablation studies of CS/SS feature channels for unsupervised arbitrary style transfer (day $\rightarrow$ night). A lower FID and a higher IS indicate better performance.}
\begin{tabularx}{\textwidth}{l|*{3}{|Y}}
\Xhline{1pt}
& \multicolumn{3}{c}{arbitrary style transfer (day $\rightarrow$ night)} \\
\cline{2-4}
Metrics & full model& $\mathrm{channels}\div2$& $\mathrm{channels}\div4$ \\
\Xhline{0.7pt}
FID $\downarrow$& {\bf79.697}& 82.357& 95.199 \\
IS $\uparrow$& {\bf2.203}& 2.142& 2.101 \\
\Xhline{1pt}
\end{tabularx}
\label{tbl:ablation_channel_ast}		
\end{table}

\begin{table}[tb!]
\centering
\caption{The quantitative evaluation on ablation studies of CS/SS feature channels for supervised semantic image synthesis (Cityscapes). A higher mIoU, a higher pixel accuracy (accu) and a lower FID indicate better performance.}
\begin{tabularx}{\textwidth}{l|*{3}{|Y}}
\Xhline{1pt}
& \multicolumn{3}{c}{semantic image synthesis (Cityscapes)} \\
\cline{2-4}
Metrics & full model& $\mathrm{channels}\div2$& $\mathrm{channels}\div4$ \\
\Xhline{0.7pt}
mIoU $\uparrow$& {\bf65.9}& 61.0& 56.6 \\
accu $\uparrow$& {\bf82.7}& 82.1& 81.5 \\
FID $\downarrow$& {\bf59.2}& 71.8& 74.4 \\
\Xhline{1pt}
\end{tabularx}
\label{tbl:ablation_channel_sis}		
\end{table}

\begin{table}[tb!]
\centering
\caption{The quantitative evaluation on ablation studies of CS/SS feature channels for multi-modal image synthesis. A lower FID and a higher IS indicate better performance.}
\begin{tabularx}{\textwidth}{l|*{3}{|Y}}
\Xhline{1pt}
& \multicolumn{3}{c}{multi-modal image synthesis} \\
\cline{2-4}
Metrics & full model& $\mathrm{channels}\div2$& $\mathrm{channels}\div4$ \\
\Xhline{0.7pt}
FID $\downarrow$& {\bf85.876}& 93.258& 97.297 \\
IS $\uparrow$& {\bf2.934}& 2.851& 2.813 \\
\Xhline{1pt}
\end{tabularx}
\label{tbl:ablation_channel_mmis}		
\end{table}

\vspace{0.1cm}
\noindent
\textbf{Feature-level/Image-level injection ablation studies.}
To verify the importance of the feature-level injection, We further conduct feature-level/image-level injection ablation studies.
TSIT performs feature-level injections from the content/style stream to the generator to adapt to diverse tasks.
In comparison, the direct injection of resized images (\ie, the direct application of AdaIN in arbitrary style transfer, and SPADE in semantic image synthesis) can be regarded as the image-level injections.
We provide quantitative evaluation results under this setting.
As shown in Table~\ref{tbl:ablation_injection_ast} and  Table~\ref{tbl:ablation_injection_sis}, compared to our feature-level injection scheme, the image-level injection leads to a performance drop.
This suggests the significance of feature-level injection in TSIT.

\begin{table}[tb!]
\centering
\caption{The quantitative evaluation on ablation studies of feature-level (FAdaIN)/ image-level (AdaIN) injection for unsupervised arbitrary style transfer (day $\rightarrow$ night). A lower FID and a higher IS indicate better performance.}
\begin{tabularx}{\textwidth}{l|*{2}{|Y}}
\Xhline{1pt}
& \multicolumn{2}{c}{arbitrary style transfer (day $\rightarrow$ night)} \\
\cline{2-3}
Metrics & feature-level& image-level \\
\Xhline{0.7pt}
FID $\downarrow$& {\bf79.697}& 80.618 \\
IS $\uparrow$& {\bf2.203}& 2.182 \\
\Xhline{1pt}
\end{tabularx}
\label{tbl:ablation_injection_ast}		
\end{table}

\begin{table}[tb!]
\centering
\caption{The quantitative evaluation on ablation studies of feature-level (FADE)/ image-level (SPADE) injection for supervised semantic image synthesis (Cityscapes). A higher mIoU, a higher pixel accuracy (accu) and a lower FID indicate better performance.}
\begin{tabularx}{\textwidth}{l|*{2}{|Y}}
\Xhline{1pt}
& \multicolumn{2}{c}{semantic image synthesis (Cityscapes)} \\
\cline{2-3}
Metrics & feature-level& image-level \\
\Xhline{0.7pt}
mIoU $\uparrow$& {\bf65.9}& 59.7 \\
accu $\uparrow$& {\bf82.7}& 81.7 \\
FID $\downarrow$& {\bf59.2}& 60.1 \\
\Xhline{1pt}
\end{tabularx}
\label{tbl:ablation_injection_sis}		
\end{table}

\section{More Examples of Generated Images}
\label{sec:suppexamples}
\begin{figure}[t]
   \centering
       \includegraphics[width=\linewidth]{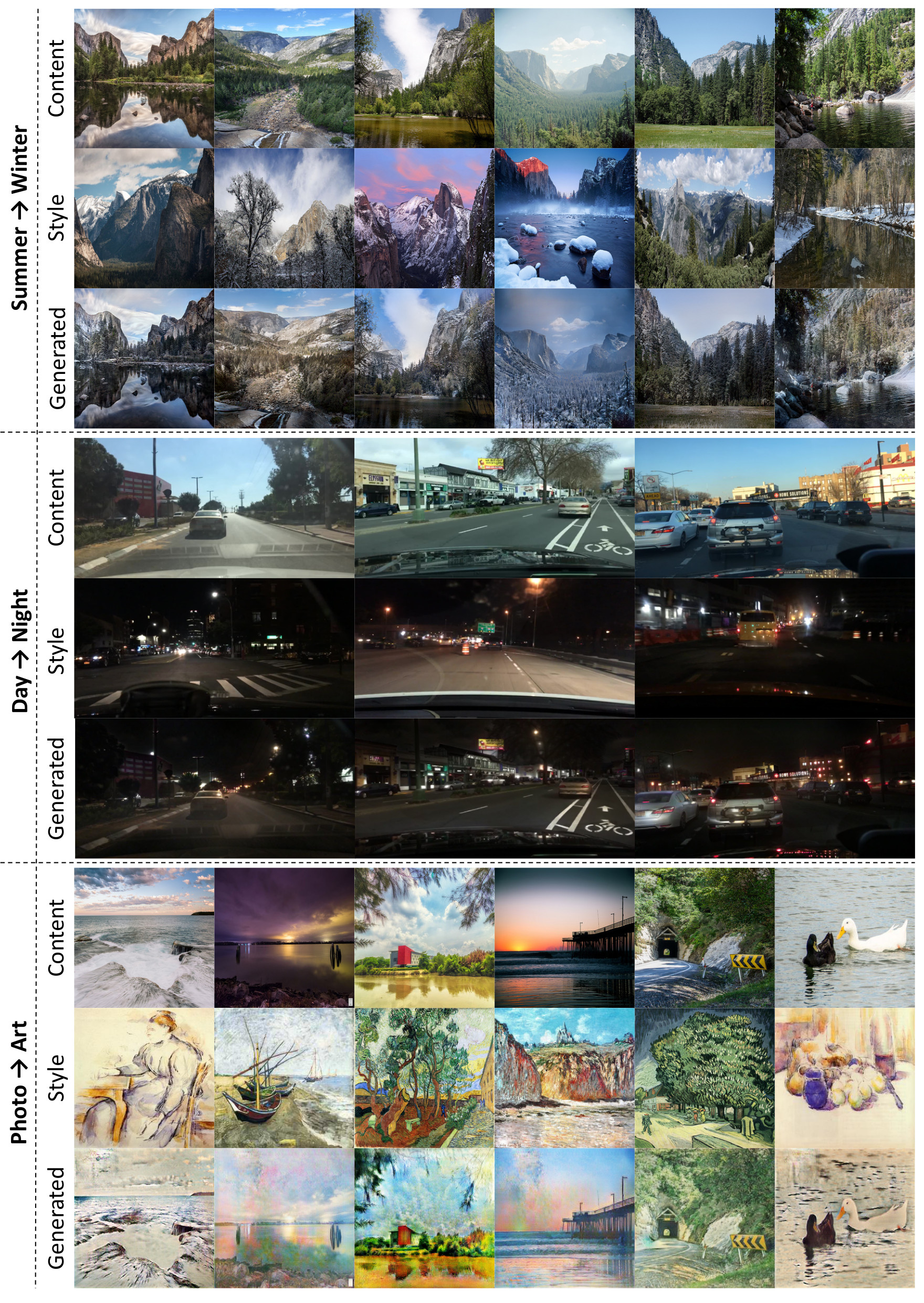}
      \caption{More examples of images generated by our method in the arbitrary style transfer task (unsupervised). Rows $1$-$3$ show Yosemite summer $\rightarrow$ winter season transfer results. Rows $4$-$6$ are BDD100K day $\rightarrow$ night translation results. Rows $7$-$9$ present photo $\rightarrow$ art style transfer results.}
   \label{fig:supp_unsup}
\end{figure}

\begin{figure}[t]
   \centering
       \includegraphics[width=\linewidth]{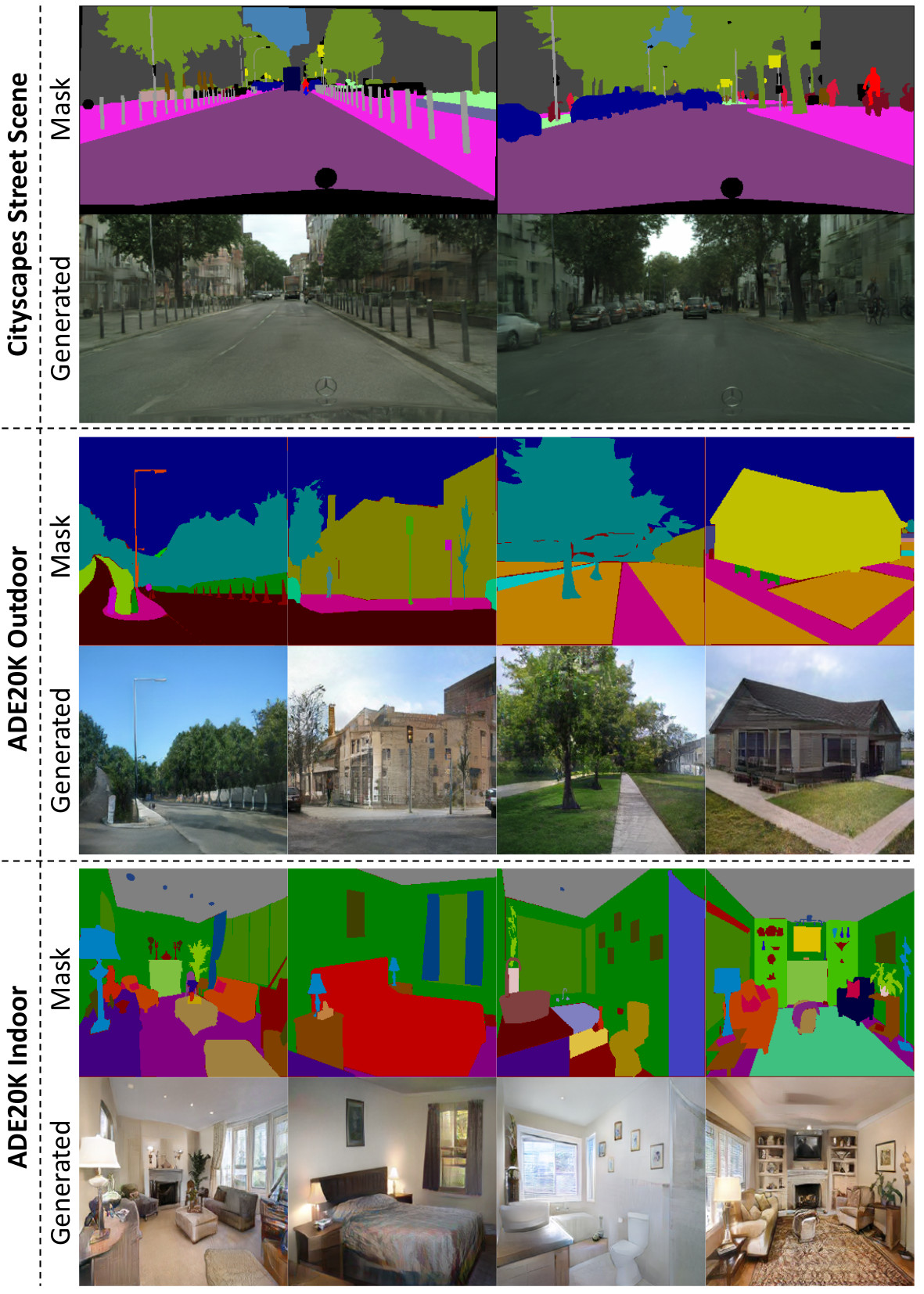}
      \caption{More examples of images generated by our method in the semantic image synthesis task (supervised). Row $1$ and $2$ show generated results on Cityscapes dataset. Row $3$ and $4$ are outdoor synthesized results on ADE20K dataset. Row $5$ and $6$ present indoor synthesized results on ADE20K dataset.}
   \label{fig:supp_sup}
\end{figure}

We show more examples of generated results by our method in Fig.~\ref{fig:supp_unsup} and Fig.~\ref{fig:supp_sup}. 
%
Several generated images of arbitrary style transfer, covering diverse scenarios, are presented in Fig.~\ref{fig:supp_unsup}.
%
We also show more synthesized exmaples of semantic image synthesis in Fig.~\ref{fig:supp_sup}. These examples feature both outdoor and indoor scenes, generated from the corresponding semantic segmentation label maps.
All the images synthesized by our proposed method are very photorealistic.

\end{document}